\documentclass[10pt,twocolumn,letterpaper]{article}

\usepackage{iccv}
\usepackage{times}
\usepackage{epsfig}
\usepackage{graphicx}
\usepackage{amsmath}
\usepackage{amssymb}
\usepackage{booktabs}
\usepackage{multirow}
\usepackage[symbol]{footmisc}

\usepackage[pagebackref=true,breaklinks=true,letterpaper=true,colorlinks,bookmarks=false]{hyperref}

\iccvfinalcopy 

\def\iccvPaperID{6304} 

\ificcvfinal\fi

\begin{document}

\title{Retro-FPN: Retrospective Feature Pyramid Network for Point Cloud Semantic Segmentation}

\author{Peng Xiang\textsuperscript{1}\footnotemark[1], \hspace{2mm}Xin Wen\textsuperscript{2}\footnotemark[1], \hspace{2mm}Yu-Shen Liu\textsuperscript{1}\footnotemark[2], \hspace{2mm}Hui Zhang\textsuperscript{1}\footnotemark[2], \hspace{2mm}Yi Fang\textsuperscript{3}, \hspace{2mm}Zhizhong Han\textsuperscript{4}\\
\textsuperscript{1}School of Software, Tsinghua University, Beijing, China\\
\textsuperscript{2}JD.com, Beijing, China\hspace{3mm}\textsuperscript{3}New York University Abu Dhabi\hspace{3mm}\textsuperscript{4}Wayne State University\\
{\small xiangp23@mails.tsinghua.edu.cn\hspace{3mm}wenxin16@jd.com\hspace{3mm}liuyushen@tsinghua.edu.cn}\\
{\small huizhang@tsinghua.edu.cn\hspace{3mm}yfang@nyu.edu\hspace{3mm}h312h@wayne.edu}
}

\begin{figure}[t]

\twocolumn[{%
\maketitle
\begin{center}
    \centering
   \includegraphics[width=\textwidth]{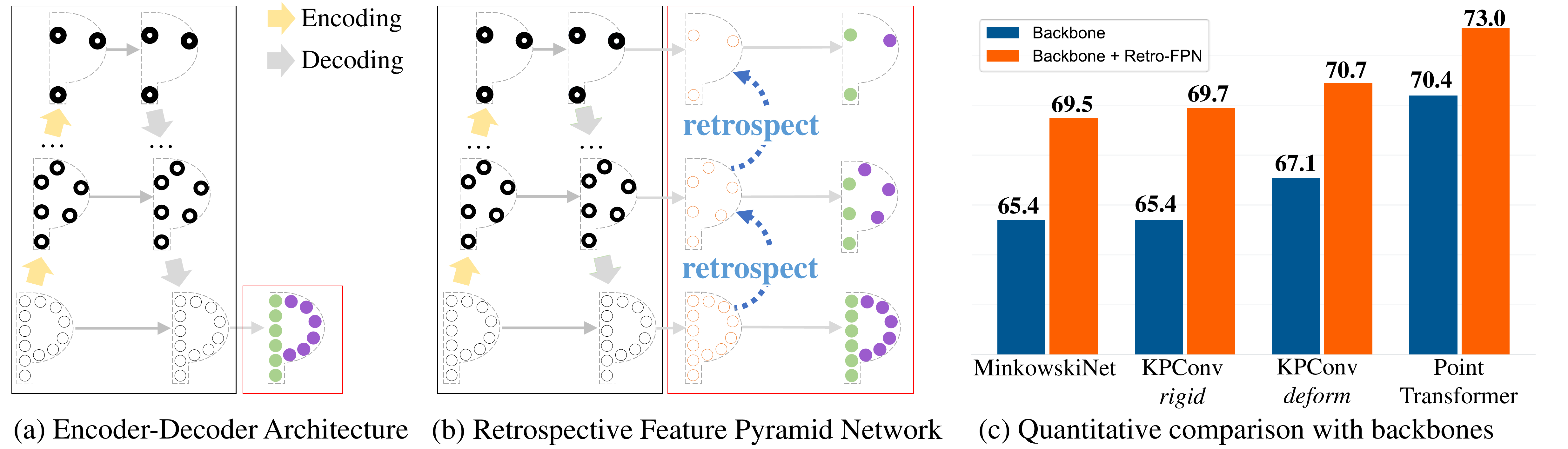}
   \caption{(a) Encoder-decoder architecture with an inherent feature pyramid in the decoding stage. Black points with thicker outlines denote region features of larger local regions, green and purple points are the predicted semantic labels. (b) Retrospective Feature Pyramid Network, the points with orange outlines denote point-level semantic features. The rectangular areas highlighted in black and red denote local region feature learning and point-level semantic feature learning, respectively. In Retro-FPN, region information flows into points at all levels, and are retrospectively refined to the lowest level. (c) mIoU on S3DIS Area 5 with and without Retro-FPN.}
   \label{fig:barchart}
\end{center}
}]
\end{figure}

\renewcommand{\thefootnote}{\fnsymbol{footnote}}
\footnotetext[1]{Equal contribution.}
\footnotetext[2]{Corresponding authors. This work was supported by National Key R\&D Program of China (2022YFC3800600), the National Natural Science Foundation of China (62272263, 62072268), and in part by Tsinghua-Kuaishou Institute of Future Media Data.}

\ificcvfinal\thispagestyle{empty}\fi

\begin{abstract}
\vspace{-2mm}
   Learning per-point semantic features from the hierarchical feature pyramid is essential for point cloud semantic segmentation. However, most previous methods suffered from ambiguous region features or failed to refine per-point features effectively, which leads to information loss and ambiguous semantic identification. To resolve this, we propose Retro-FPN to model the per-point feature prediction as an explicit and retrospective refining process, which goes through all the pyramid layers to extract semantic features explicitly for each point. Its key novelty is a retro-transformer for summarizing semantic contexts from the previous layer and accordingly refining the features in the current stage. In this way, the categorization of each point is conditioned on its local semantic pattern. Specifically, the retro-transformer consists of a local cross-attention block and a semantic gate unit. The cross-attention serves to summarize the semantic pattern retrospectively from the previous layer. And the gate unit carefully incorporates the summarized contexts and refines the current semantic features. Retro-FPN is a pluggable neural network that applies to hierarchical decoders. By integrating Retro-FPN with three representative backbones, including both point-based and voxel-based methods, we show that Retro-FPN can significantly improve performance over state-of-the-art backbones. Comprehensive experiments on widely used benchmarks can justify the effectiveness of our design. The source is available at \url{https://github.com/AllenXiangX/Retro-FPN}.
\end{abstract}

\vspace{-4mm}
\section{Introduction}
\vspace{-1mm}
3D point cloud semantic segmentation \cite{kirillov2019panoptic, JointPointBased, PointASNL, seggcn, cortinhal2020salsanext, qiu2022gfnet, PolarNet, tangentconv, xu2020squeezesegv3}, which aims to predict a unique category label for each point, is a critical task towards the 3D visual understanding of large-scale scenes. A typical solution to predict per-point semantic labels is the widely used encoder-decoder framework \cite{hu2020randla}. The encoder aims to learn contextual region features by gradually enlarging receptive fields. The decoder propagates the local region features from the larger receptive fields into the smaller ones, which inherently forms a feature pyramid \cite{lin2017feature} (see Figure \ref{fig:barchart} (a)).

\begin{figure}[t]
   \centering
   \includegraphics[width=\linewidth]{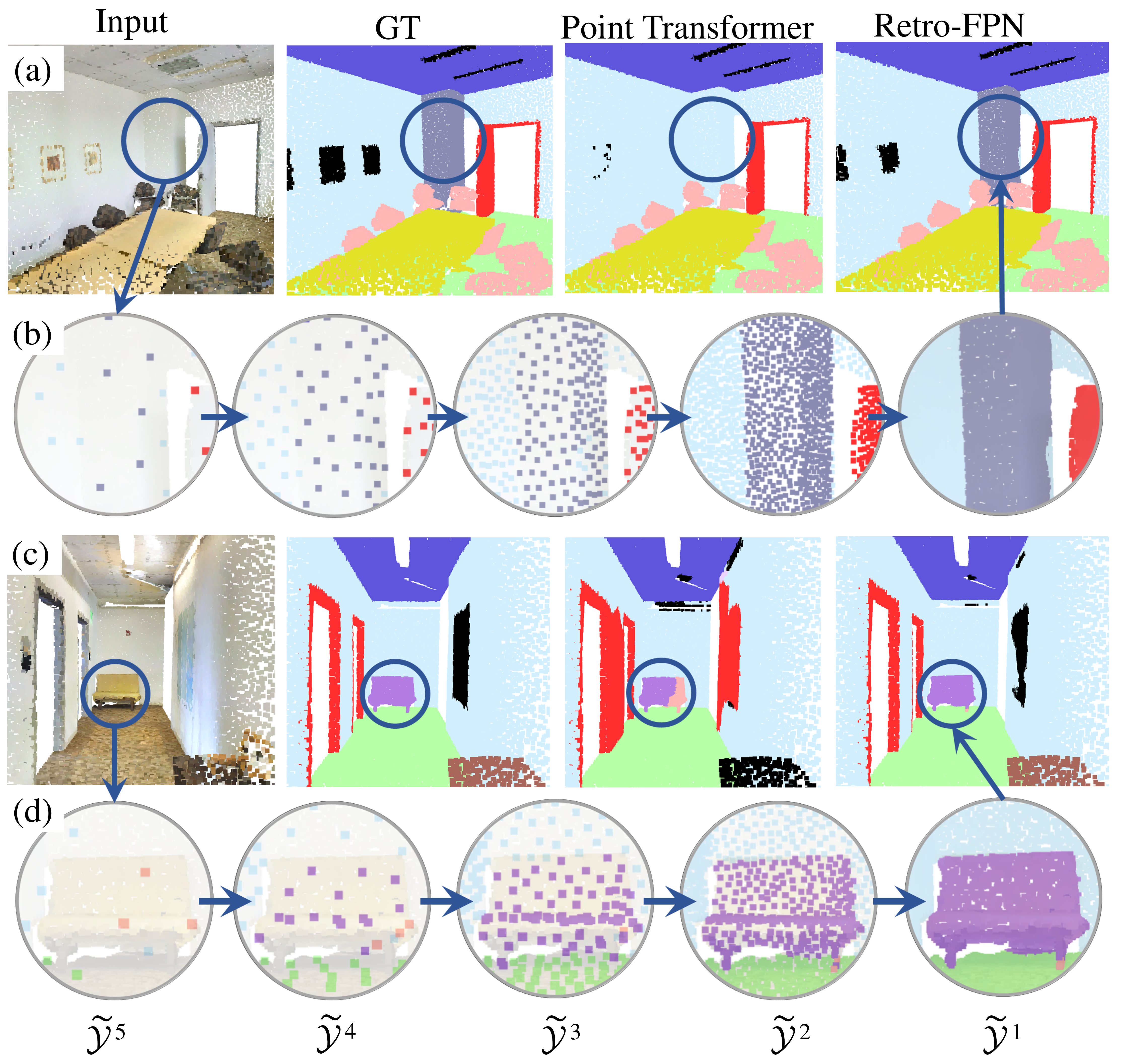}
   \caption{Visualization of segmentation process of Retro-FPN. (a) and (c) show the visual comparison with the backbone (Point Transformer \cite{zhao2021point}) network. In (a), the backbone loses the information of the column. In (c), the backbone struggles to distinguish between chair and sofa. In (b) and (d), we show the retrospective refining process by Retro-FPN over the improved areas.
   }
   \label{fig:teaser}
\end{figure}

Learning per-point feature prediction from the pyramidal region features is the target of point cloud semantic segmentation. However, most existing encoder-decoder-based networks merely reveal per-point features explicitly at the final layer (denoted as red box in Figure \ref{fig:barchart} (a)), leaving abundant semantic information stuck in the intermediate region features (black box in Figure \ref{fig:barchart} (a)), which cannot directly facilitate the final prediction. This may lead to the loss of semantic information and ambiguous semantic identification, as demonstrated in Figure \ref{fig:teaser} (a) and (c). Since each pyramid layer may contain useful and erroneous information simultaneously, the per-point semantic features should be carefully refined through all stages. 

To resolve this, some prior works \cite{gong2021omni, Kirillov_2020_CVPR} adopt hierarchical supervision to refine intermediate predictions explicitly. In 2D vision, PointRend \cite{Kirillov_2020_CVPR} proposed to refine high-frequency points with hierarchical supervision, but each point is refined based on the features interpolated at a single location, which suffered to capture the local semantic pattern and may fail to obtain informative per-point features for 3D point clouds. RFCR \cite{gong2021omni} first introduced multi-scale supervision to point cloud semantic segmentation, but the supervision was on region-level and it's still difficult to obtain accurate per-point prediction from the region features.


Therefore, we propose Retro-FPN to improve per-point  semantic feature prediction by fully utilizing the feature pyramid, which is achieved by an explicit and retrospective refining process (see Figure \ref{fig:barchart} (b)). Specifically, by predicting per-point labels for all the middle layers, Retro-FPN allows region information to flow into points and obtains the point-level semantic features at each stage. Then, the features are carefully refined by retrospectively summarizing the semantic pattern from the previous layer and adaptively rearranging the current semantic information. 


To conduct retrospective refinement, we introduce a novel \emph{retro-transformer} in each layer to extract per-point semantic features, which consist of two stages. The first stage aims to ``retrospect'' useful information from the previous layer. Since the category of each point is similar to its surrounding local region, we use a local cross-attention block to conduct retrospection, which takes the features of the current layer as queries to summarize semantic contexts from the previous layer. Different from the region-level information in the backbone features, such contextual information are built upon the per-point semantic features of the nearby points, which can fully facilitate the refinement of each point by selectively revisiting its neighbor points. The second stage serves to ``refine'' the current semantic features by combining them with the summarized contexts. Instead of merging the features with simple adding or concatenation, we use a lightweight semantic gate to adaptively preserve and forget the previous semantic information. The retro-transformer can establish a cross-level semantic relationship between different decoding stages, this enables the network to explicitly preserve useful information and discard erroneous information in each stage, as illustrated in Figure \ref{fig:teaser} (b) and (d).

Retro-FPN is a pluggable neural network that can extract and refine per-point semantic features for prevailing backbones, including both point-based and voxel-based methods. Specifically, we embed Retro-FPN into KPConv \cite{thomas2019kpconv}, MinkowskiNet \cite{Choy_2019_CVPR}, Point Transformer \cite{zhao2021point}, and Point Transformer V2\cite{wu2022pointv2}. Non-trivial improvements on the S3DIS \cite{s3dis} Area 5 benchmark (Figure \ref{fig:barchart} (c)) can verify the effectiveness of our network design. In summary, our contributions are threefold:
\begin{itemize}
\item We propose Retro-FPN to improve per-point semantic feature prediction for 3D point clouds. Retro-FPN models the feature propagation as an explicit and retrospective refining process on point-level semantic information, which is a plug-and-play network that can improve the performance of prevailing backbones. 
\item We propose a novel retro-transformer to establish a cross-level semantic relationship between different decoding stages. It utilizes a local cross-attention to retrospect the previous semantic pattern and leverages a lightweight semantic gate unit to refine the current semantic features.
\item We integrate Retro-FPN with both point-based and voxel-based backbones and evaluate our method on the S3DIS \cite{s3dis}, ScanNet \cite{dai2017scannet} and SemanticKITTI \cite{semantickitti} benchmarks. Experimental results demonstrate that our method can significantly improve performance over state-of-the-art methods.

\end{itemize}

\section{Related Work}

\noindent\textbf{Point cloud semantic segmentation}. In recent years, the tremendous development of deep learning-based \cite{ma2023towards, zhou2023levelset, zhou2022learning, wen20223d, zhang2022fast, hzz2023qq} 3D processing techniques \cite{li2023shsnet, liu2022spu, liu2021fine} has significantly boosted the progress of point cloud semantic segmentation \cite{huang2018recurrent, spgraph, zhang2019shellnet, Peng2023OpenScene, ding2022language}, which can be roughly divided into two categories. (1) The point-based \cite{pointnet, wu2019pointconv, paconv, thomas2019kpconv, zhao2021point, pointweb, segcloud_3dv2017, 3DMV} methods directly handle raw point clouds. As one of the pioneering works, Point-Net++ \cite{qi2017pointnet++} used a local sampling and grouping mechanism to extract contextual information. Followers along this line focus on effective feature aggregation technique to obtain representative features, such as convolution-like operations \cite{thomas2019kpconv, pointcnn} and the attention mechanism \cite{vit, zhao2021point, lai2022stratified, fastpointtransformer, wu2022pointv2}. (2) The voxel-based \cite{Choy_2019_CVPR, Graham_2018_CVPR_sparse} methods first transform 3D point clouds into voxels, then apply sparse convolutions to learn point cloud representations. While these methods can handle large-scale scenes, they also suffer from detailed information loss due to voxelization.  For both point-based and voxel-based methods, an encoder-decoder architecture is a typical solution. While previous methods \cite{wu2019pointconv, paconv} usually highlight the importance on feature aggregation in the encoding stage, we concentrate on the explicit decoding of semantic information to unleash the performance for prevailing backbones.

\noindent\textbf{Pyramidal feature representation}. The feature pyramid is an important component of deep neural networks, which can perceive large-scale scenes at different scales. FPN \cite{lin2017feature} is a pioneering work that leverages the pyramid features to detect multi-scale objects. Since then, the feature pyramid has been explored in 2D dense prediction tasks, such as object detection \cite{NAS_FPN_2019_CVPR, effidet_Tan_2020_CVPR}, instance segmentation \cite{liu2018path, SSAP_2019_ICCV, A2FPN_2021_CVPR} and panoptic segmentation \cite{kirillov2019panoptic}. Semantic segmentation requires per-point prediction at the final layer, to exploit the feature pyramid, one possible solution is to up-sample intermediate features \cite{Lin_2017_CVPR, Nie_2022_CVPR} or predictions to the finest resolution and fuse them like BAAF-Net \cite{BAAF_Net_Qiu_2021_CVPR} and PANet \cite{liu2018path}. However, each pyramid layer may contain useful and erroneous information simultaneously, simply fusing the intermediate outputs can lead to false predictions. Another solution is to incorporate hierarchical supervision and refine the intermediate predictions by layer. In 2D vision, PointRend \cite{Kirillov_2020_CVPR} proposed to gradually refine points in high-frequency areas, but each point is refined based on the interpolated prediction and features at a single location, which cannot provide adequate local contexts for refinement. Furthermore, the point selection procedure of PointRend is tailored for dense and regular 2D grids, which cannot directly apply to point cloud data. RFCR \cite{gong2021omni} is one of the first attempts to utilize feature pyramid with hierarchical supervision for 3D point clouds, but it focused merely on enhancing region level semantic features, which is difficult to fully preserve and refine per-point semantic information at each stage.

Compared with the previous methods, Retro-FPN takes a step further to explore a context-aware solution for refining semantic features on per-point level, which is tailored for 3D point clouds. Retro-FPN refines each point based the local semantic contexts retrospected from the previous layer and selectively preserve and forgo semantic information in consecutive layers, which enables to fully unleash the potential of prevailing backbones.

\noindent\textbf{Relation to transformer}. 
Transformer \cite{vit} was first proposed for natural language processing and soon became dominant in 2D computer vision \cite{liu2021Swin}. Inspired by this success, many studies \cite{zhao2021point, PCT_Guo_2021, pointformer_Pan_2021_CVPR, xiang2023SPD, xiang2021snowflakenet} have attempted to leverage the representation ability of transformer to process 3D point clouds \cite{pmpnet++, wen2021pmp, li2023neaf, BaoruiNoise2NoiseMapping, NeuralTPS, wen2020point2spatialcapsule}. Recently, More studies further explored the attention mechanism that caters to point clouds, including the study of long range dependency \cite{lai2022stratified}, efficient attention mechanism \cite{fastpointtransformer} and powerful local attention \cite{wu2022pointv2}. While these methods have made substantial progress, they use self-attention for representation learning in a single stage. Differently, we propose retro-transformer to establish semantic relationships across different decoding stages.

\noindent\textbf{Plug-and-play network}. Plug-and-play networks \cite{lu2021cga, qiu2021pnp-3d, gong2021omni} aim to benefit multiple backbones as a plug-in module. They become imperative as recent advanced 3D semantic segmentation backbones was introduced. For example, CGA-Net \cite{lu2021cga} addresses feature augmentation with inter and intra-class consistency. PnP-3D \cite{qiu2021pnp-3d} targets the local-global feature fusion. RFCR \cite{gong2021omni} enhances region-level backbone features with omni-supervision. Different from the above methods, Retro-FPN explores the per-point level semantic prediction, which can bring improvement for hierarchical decoders 
 \cite{thomas2019kpconv, zhao2021point, Choy_2019_CVPR, wu2022pointv2} including both point-based and voxel-based backbones.

\begin{figure*}[!t]
   \centering
   \includegraphics[width=\textwidth]{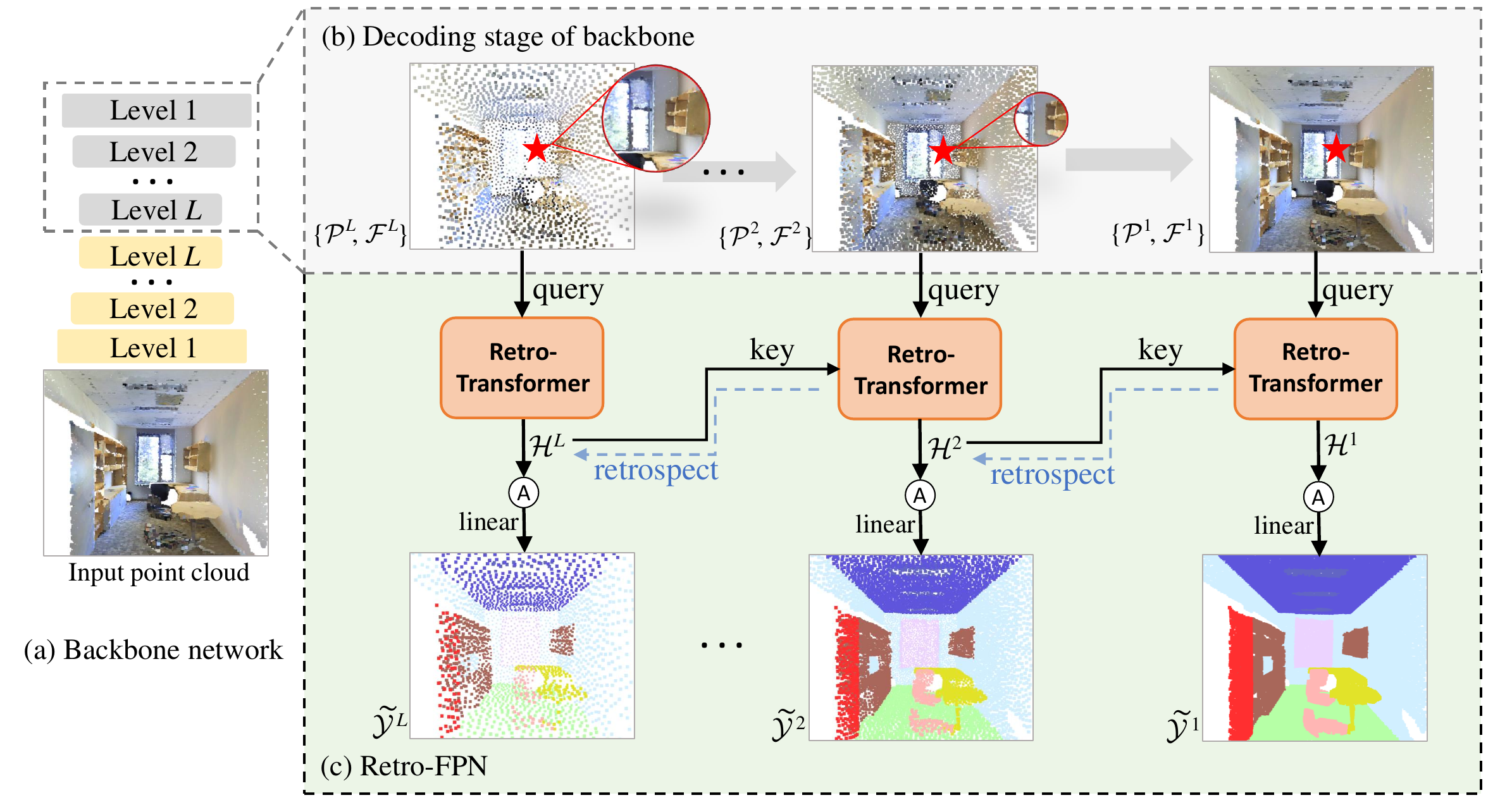}
   \caption{(a) shows an encoder-decoder architecture. (b) In the decoding stage of backbone, only three pyramid layers (1, 2 and $L$) are shown for clarity, $\mathcal{P}^l$ is point set in each decoding stage, and $\mathcal{F}^l$ is the region feature of $\mathcal{P}^l$. The larger circular area highlighted in red denotes larger local region around $\mathcal{P}^l$, which is characterized by $\mathcal{F}^l$. (c) For Retro-FPN, $\mathcal{H}^{l+1}$ is point-level semantic feature from previous layer, which provides key and value for retro-transformer. $\mathcal{F}^l$ provides query to retrospectively summarize semantic pattern from $\mathcal{H}^{l+1}$.}
   \label{fig:overall}
\end{figure*}

\section{Method}
\subsection{Overview and Motivation}
We show a typical encoder-decoder architecture with $L$ levels in Figure \ref{fig:overall} (a), and the inherent feature pyramid hierarchy of the decoding stage is shown in Figure \ref{fig:overall} (b). Our Retro-FPN is integrated with the backbone decoder and shown in Figure \ref{fig:overall} (c). For clarity, we only visualize three pyramid layers (1, 2 and $L$).

As shown in Figure \ref{fig:overall} (b), we denote the point set in each decoding stage as $\mathcal{P}^l \in \mathbb{R}^{N_l \times 3}$, and the local context around $\mathcal{P}^l$ is denoted as the region feature $\mathcal{F}^l \in \mathbb{R}^{N_l \times C_l}$. From $\mathcal{P}^L$ to $\mathcal{P}^1$, the decoder propagates contextual information from the larger receptive (highlighted in the red circle) fields into the smaller ones, and finally to the point-level features $\mathcal{F}^1$. However, there are two problems with this paradigm. First, the backbone decoder propagates semantic information simplicitly, where the long path from the intermediate levels (layer 2-$L$) to the prediction layer (layer 1) may cause information loss.
Second, although the high-level features have large receptive fields, it is still difficult to precisely capture the accurate semantic contexts of the underlying local regions, especially when there are different semantic objects within the same region, e.g., at the boundary of window, wall and bookcase.

Based on the above observation, we propose Retro-FPN extract accurate per-point semantic features from the feature pyramid, which is conducted by explicitly and retrospectively refining the point-level semantic information.

\subsection{Retro-FPN}
As shown in Figure \ref{fig:overall} (c), Retro-FPN is designed to explicitly extract and refine semantic information for all pyramid levels. In level $l$, the region feature $\mathcal{F}^l$ is first refined and converted into point-level semantic feature $\mathcal{H}^l$ by a retro-transformer. Then, we explicitly predict per-point labels $\tilde{\mathcal{Y}}^l$ from $\mathcal{H}^l$ using an activation function followed by a linear transformation. 

There are two advantages to the design of Retro-FPN. First, instead of struggling to perceive the complex local regions like RFCR \cite{gong2021omni}, the explicit prediction of per-point labels allows Retro-FPN to focus on point level semantic information. The intuition is that for a point $\mathbf{p}_i \in \mathcal{P}^l$, it is easier to identify its single semantic category than recognize all the semantic objects within the surrounding local region. This scheme enables Retro-FPN to incorporate accurate semantic information into $\mathcal{H}^l$, which significantly facilitates the retrospective refinement. Second, although the global contexts are essential for scene understanding, the saturated contextual prior could hamper the network to perceive detailed local semantic information \cite{nekrasov2021mix3d}. This problem could be even worse in higher pyramid layers. Hence, encouraging the middle layers to focus on per-point semantic information can help the network to balance global scene contexts and the detailed semantic information.

While the overall architecture of Retro-FPN can help to learn accurate per-point semantic information, now the more critical problem is to refine the information and facilitate the final prediction. Since $\mathcal{H}^l$ from intermediate layers ($l > 1$) may contain false semantic information, two goals have to be achieved: (1) preserving useful information and (2) discarding erroneous information. Previous methods like PointRend \cite{Kirillov_2020_CVPR} refine each point based on the coarse prediction and the interpolated features at a single location, but the accurate semantic category of each point is dominated by its local neighborhood, the interpolated features cannot provide adequate semantic contexts for per-point refinement. Differently, we propose a novel \emph{retro-transformer} and leverage attention mechanism to selectively summarize local semantic information. The detailed structure of retro-transformer is described below.



\begin{figure}[t]
   \centering
   \includegraphics[width=\linewidth]{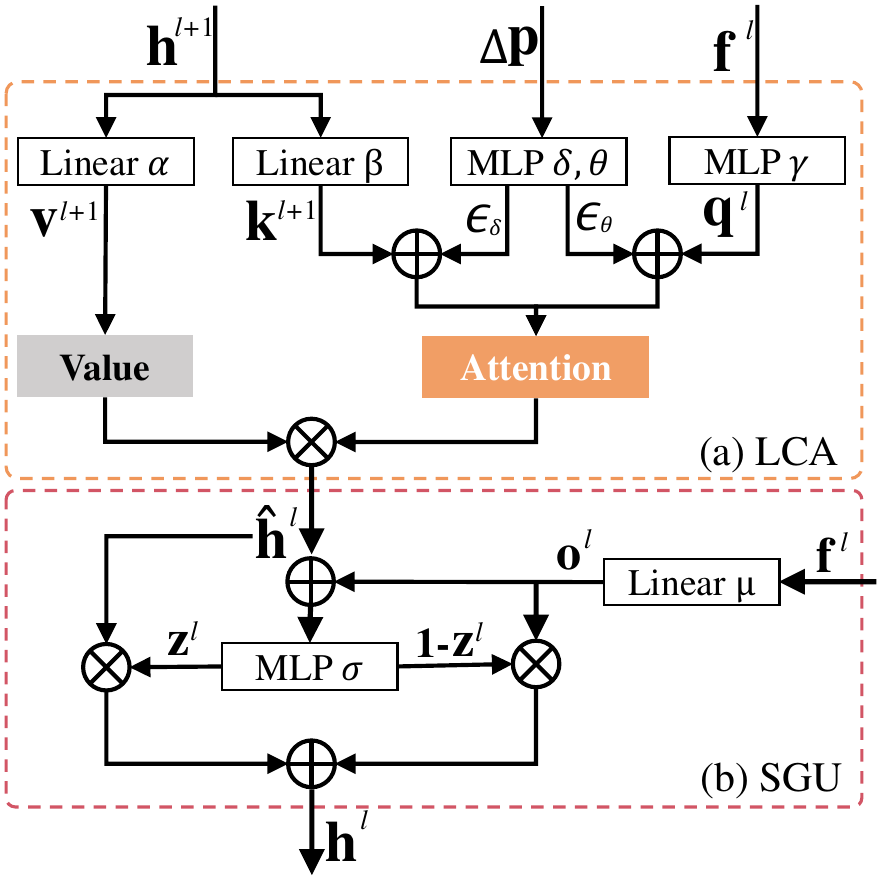}
   \caption{The structure of Retro-Transformer. (a) The local cross-attention (LCA) block. (b) The semantic gate unit (SGU).}
   \label{fig:retro_transformer}
   \vspace{-1em}
\end{figure}

\subsection{Retro-Transformer}
\label{section:retro_transformer}

The structure of Retro-Transformer is shown in Figure \ref{fig:retro_transformer}, which consists of a local cross-attention block (Figure \ref{fig:retro_transformer} (a)) and a semantic gate unit (Figure \ref{fig:retro_transformer} (b)). The cross-attention aims to conduct ``retrospection''. The per-point semantic features from the previous layer can provide rich semantic contexts and guide the current layer. Hence, for each point, we leverage the attention mechanism to attentively summarize semantic contexts by revisiting its neighbor points from the previous layer. Further, The semantic gate serves to achieve ``refinement''. Because intermediate semantic features will inevitably contain erroneous information, the gate mechanism allows the retro-transformer to selectively retain and forgo information from both the previous and the current layer.




\noindent\textbf{Local cross-attention block}. As shown in Figure \ref{fig:retro_transformer} (a), the cross-attention takes the previous semantic feature $\mathbf{h}^{l+1} \in \mathbb{R}^C$ and the current region feature $\mathbf{f}^l \in \mathbb{R}^C_l$ as inputs to summarize semantic contexts. Since $\mathbf{f}^l$ and $\mathbf{h}^{l+1}$ are from different branches and may have large discrepancy, unlike previous transformers \cite{vit, zhao2021point} that produce the query vector with a linear layer, we use the non-linear transformation of multi-layer perceptron (MLP) to obtain $\mathbf{q}^l \in \mathbb{R}^C$, which can bridge the gap between the two branches with more learnable capacities. Then, the value and key vectors are produced from $\mathbf{h}^{l+1}$ using linear layer as follows:
\begin{equation}
    \begin{split}
    \label{eq:qkv}
    \mathbf{q}^l = & \mathrm{MLP}_\gamma (\mathbf{f}^l),  \\   
    \mathbf{v}^{l+1} = Linear_\alpha (\mathbf{h}^{l+1}), & \quad
    \mathbf{k}^{l+1} = Linear_\beta (\mathbf{h}^{l+1}).
    \end{split}
\end{equation}
Furthermore, since the semantic information of each point $\mathbf{p}_i^l$ is dominated by the surrounding local region, we adopt local attention to aggregate semantic contexts from its nearby points $\mathbf{p}_j^{l+1}$ (subscript $i$ and $j$ denote the point index) in the previous layer. The neighborhood of $\mathbf{p}_i^l$ is defined as the $\mathrm{K}$-nearest neighbor ($\mathrm{K}$-NN) points. The K-NN strategy lets retro-transformer focus on local semantic contexts, which also reduces computation cost significantly. It is worth noting that the point clouds of the previous layer are usually much sparser than the current ones, so that even a small $\mathrm{K}$-NN search can effectively enlarge receptive field. Moreover, since the complex local region may increase the difficulty for learning robust contexts, we enhance the query and key vectors with learnable position embedding to incorporate positional relationship. Specifically, for each $q_i^l$, we denote the key vectors of the K-nearest neighbors as $\{ \mathbf{k}^{l+1}_{i, k} | k=1,2,\dots,\mathrm{K} \}$, where subscript $k$ denotes the $k$-th neighbor and calculate attention as:
\begin{equation}
    \label{eq:attention}
    w_{ik} = \langle \mathbf{q}_i^l + \epsilon_\delta, \, \mathbf{k}_{i, k}^{l+1} + \epsilon_\theta \rangle \\/ \sqrt{C},
\end{equation}
where the position embedding $\epsilon_\delta$ and $\epsilon_\theta$ are obtained by passing the relative position $\Delta \mathbf{p}$ ($\Delta \mathbf{p} = \mathbf{p}_i^l - \mathbf{p}_{i, k}^{l+1}$) through two MLPs. Then, the aggregated semantic contexts $\mathbf{\hat{h}}_i^l$ are given as follows:
\begin{equation}
    \label{eq:h_hat}
    \mathbf{\hat{h}}_i^l = \sum\limits_{k=1}^\mathrm{K} Softmax(\mathbf{w}_i)_k\mathbf{v}_{i, k}.
\end{equation}
Note that there is no $\mathbf{h}^{L+1}$ for the highest pyramid layer (the $L$-th layer), where the cross-attention degrades to self-attention and takes $\mathbf{f}^L$ as query, key and value.

\noindent\textbf{Semantic gate unit.} As shown in Figure \ref{fig:retro_transformer} (b), we refine the region feature $\mathbf{f}^l$ with the summarized semantic contextual feature $\mathbf{\hat{h}}^l$ using the gate mechanism. To reduce computation cost, we take inspiration from gated recurrent unit (GRU) \cite{chung2014empirical} and adopts a single update gate to control information flow. Specifically, given region feature $\mathbf{f}^l \in \mathbb{R}^C_l$, we first compacts its information into vector $\mathbf{o}^l \in \mathbb{R}^l$ by $\mathbf{o}^l = Linear_\mu (\mathbf{f}^l)$. Then, the update gate $\mathbf{z}^l$ is given as:
\begin{equation}
    \label{eq:update_gate}
    \mathbf{z}^l = \mathrm{MLP}_\sigma (\mathbf{\hat{h}}^l + \mathbf{o}^l).
\end{equation}
Finally, we obtain the point-level semantic feature $\mathbf{h}^l$ by the following equation:
\begin{equation}
    \label{eq:h}
    \mathbf{h}^l = \mathbf{z}^l \odot \mathbf{\hat{h}}^l + (1-\mathbf{z}^l) \odot \mathbf{o}^l.
\end{equation}
\subsection{Integration with backbones}
\label{section:voxel_point_based_backbones}
Retro-FPN can be integrated with prevailing backbones that adopt an encoder-decoder architecture, including both point-based and voxel-based methods. To employ Retro-FPN, we only need the point set $\mathcal{P}^l$ of each decoding stage, the corresponding region feature $\mathcal{F}^l$ and the ground-truth label $\mathcal{Y}^l$. For point-based methods, we record the ground-truth labels $\mathcal{Y}^l$ along the downsampling process of the encoding stage, and directly use $\mathcal{F}^l$ from the decoder. For voxel-based methods, we take the voxels in each layer as intermediate point clouds and also focus on learning per-point semantic information from the voxel features. Since each voxel may correspond to multiple category labels, we use the most common one as its ground-truth label. Moreover, for both point-based and voxel-based backbones, the intermediate layer may contain too many points (voxels) due to small downsampling rates, which severely increases computation cost. Meanwhile, the $\mathrm{K}$-NN search in a dense point cloud also leads to limited receptive fields. To avoid the above problems, we further use random sampling to downsample the intermediate point clouds.

\begin{table}
  \caption{Quantitative results on S3DIS \cite{s3dis} dataset, evaluated on Area 5. \textcolor{red}{Red} number means better results than baseline. \textbf{Bold} numbers denote the best results among all methods. * denotes  voting augmentation during testing.}
  \label{table:s3dis_area5}
  \centering
  \begin{tabular}{l|c|c}
    \toprule
    Method & Input &mIoU\\
    \midrule
    
    CGA-Net \cite{lu2021cga} & point/voxel & 68.6\\
    PnP-Net \cite{qiu2021pnp-3d} & point & 68.5\\
    RFCR \cite{gong2021omni} & point & 68.7\\
    DeepViewAgg \cite{multiview_aggregation} & point + 2D & 67.2 \\
    RepSurf \cite{repsurf} & point & 68.9\\
    CBL \cite{Tang2022ContrastiveBL} & point & 71.0\\
    Fast Transformer \cite{fastpointtransformer} & point & 70.3 \\
    EQ-Net \cite{eqnet} & point/voxel & 71.3\\
    Stratified Transformer \cite{lai2022stratified} & point & 72.0\\
    Point Mixer \cite{choe2022pointmixer} & point & 71.4 \\
    Point Transformer V2 \cite{wu2022pointv2} & point & 71.6\\
    \midrule
    MinkowskiNet (5cm) * \cite{Choy_2019_CVPR} & voxel & 65.4\\
    MinkowskiNet + Retro-FPN * & voxel & \textcolor{red}{69.5}\\
    \midrule
    KPConv \textit{rigid} * \cite{thomas2019kpconv} & point & 65.4\\
    KPConv \textit{rigid} + Retro-FPN * & point & \textcolor{red}{69.7}\\
    \midrule
    KPConv \textit{deform} * \cite{thomas2019kpconv} & point & 67.1 \\
    KPConv \textit{deform} + Retro-FPN * & point & \textcolor{red}{70.7}\\
    \midrule
    PointTransformer \cite{zhao2021point} & point & 70.4\\
    PointTransformer + Retro-FPN & point & \textbf{\textcolor{red}{73.0}}\\
    \bottomrule
  \end{tabular}
\end{table}

\subsection{Training loss}
We use cross entropy loss to guide the predictions from all decoding stages, the training loss is formulated as $\mathbf{L} = \sum\lambda_l\mathbf{L}_l$, where $\mathbf{L}_l$ is the loss of the $l$-th layer. $\lambda_l$ is the weight to balance losses in each layer.

\section{Experiments}
\label{section:experiments}

\subsection{Datasets and metric}
\label{section:dataset}

\noindent\textbf{S3DIS}. The S3DIS \cite{s3dis} dataset comprises point clouds of 271 rooms in six areas. There are 273 million points in total, and each point is assigned a semantic label of 13 categories. Following previous methods \cite{qi2017pointnet++, segcloud_3dv2017, zhao2021point}, we evaluate our method on the Area 5 and 6-fold benchmarks.

\noindent\textbf{ScanNet v2}. The ScanNet v2 \cite{dai2017scannet} provides 1,613 indoor scans, where the train/val/test split is 1,201/312/100, respectively. The training and validation sets contain point-level annotations, and the test set is provided without ground-truth annotations.

\noindent\textbf{SemanticKITTI}. The SemanticKITTI \cite{semantickitti} dataset provides 43,552 LIDAR scans that belong to 21 sequences. The training set contains 19,130 scans from sequences 00-07 and 09-10, and the validation set has 4,071 scans from sequence 08. The testing set contains 20,351 scans from sequences 11-21, which is set for online testing and only the 3D coordinates are provided. 

\begin{table}
  \caption{Quantitative results on S3DIS \cite{s3dis} dataset, evaluated on 6-fold cross validation.}
  \label{table:s3dis_6fold}
  \centering
  \begin{tabular}{l|c}
    \toprule
    Method & mIoU\\
    \midrule
    KPConv \cite{thomas2019kpconv}  & 70.6 \\
    FPConv \cite{fpconv}  & 68.7 \\
    PAConv \cite{paconv}  & 69.3 \\
    SCF-Net \cite{SCF_Net}  & 71.6 \\
    CBL \cite{Tang2022ContrastiveBL}  & 73.1\\
    DeepViewAgg \cite{multiview_aggregation}  & 74.7 \\
    RepSurf \cite{repsurf}  & 74.3 \\
    EQ-Net \cite{eqnet}  & \textbf{77.5} \\
    PointNeXt \cite{qian2022pointnext}  & 74.9\\
     \midrule
    PointTransformer \cite{zhao2021point}  & 73.5 \\
    PointTransformer + Retro-PFN & \textcolor{red}{77.3}\\
    \bottomrule
  \end{tabular}
\end{table}

\noindent\textbf{Evaluate metric.} For the above benchmarks, we adopt the mean Intersection-over-Union (mIoU) as evaluation metric.

\subsection{Backbones and experimental settings}
\label{section:backbone}
\noindent\textbf{Backbones.} On the S3DIS \cite{s3dis} Area 5 benchmark, we embed Retro-FPN into both point-based (Point Transformer \cite{zhao2021point} and KPConv \cite{thomas2019kpconv}) and voxel-based \cite{Choy_2019_CVPR} methods to prove the generalization ability of Retro-FPN. Since the six areas of S3DIS have large discrepancies, we further choose the high-performing Point Transformer to evaluate the robustness of Retro-FPN on the S3DIS 6-fold benchmark. As for the ScanNet \cite{dai2017scannet} and SemanticKITTI \cite{semantickitti} datasets, we use MinkowskiNet as backbone, because it is a more popular choice that has been widely adopted as backbone by previous methods like BPNet \cite{hu2021bidirectional} and SPVNAS \cite{tang2020searching}. Furthermore, to verify the effectiveness of Retro-FPN with state-of-the-art backbones, we integrate Retro-FPN with the Point Transformer V2 \cite{wu2022pointv2} on ScanNet. 

\noindent\textbf{Experimental settings.} We implement Retro-FPN using PyTorch \cite{pytorch}. To have fair and solid experiments, we integrate Retro-FPN based on the official implementation of the baseline methods and keep the experimental settings the same as the backbones. We provide more experimental details in the supplementary materials.

\subsection{Quantitative results}
\label{section:quantitative}

\begin{table}[t]
  \caption{Quantitative results on ScanNet v2 \cite{dai2017scannet} in terms of mIoU. * denotes voting augmentation during testing.}
      \label{table:scannet}
      \centering
  \begin{tabular}{l|cc}
  \toprule
  Method & Val & Test\\
       \midrule

        KPConv \cite{thomas2019kpconv} * & 69.2 & 68.6 \\
        JSENet \cite{hu2020jsenet} & - & 69.9 \\
        FusionNet \cite{fusionnet} & - & 68.8 \\
        SparseConvNet \cite{Graham_2018_CVPR_sparse} & 69.3 & 72.5 \\
        BPNet \cite{hu2021bidirectional} * & 73.9 & 74.9 \\
        VMNet \cite{VMNet} & 73.3 & 74.6 \\
        StratifiedFormer \cite{lai2022stratified} & 74.3 & 74.7 \\
        EQ-Net \cite{eqnet} & 75.3 & 74.3 \\
        \midrule
        MinkowskiNet (5cm) \cite{Choy_2019_CVPR} * & 68.0 & - \\
        + Retro-FPN  *  & \textcolor{red}{70.4} & - \\
        \midrule
        MinkowskiNet (2cm) \cite{Choy_2019_CVPR}  * & 72.1 & 73.6 \\
        + Retro-FPN  * & \textcolor{red}{74.0} & \textcolor{red}{74.4}  \\
        \midrule
        Point Transformer V2 \cite{wu2022pointv2} & 75.4 & \textbf{75.2}\\
        + Retro-FPN & \textbf{\textcolor{red}{76.0}} & - \\
       \bottomrule
  \end{tabular}
\end{table}

\noindent\textbf{S3DIS Area 5.} Table \ref{table:s3dis_area5} shows the results of point cloud semantic segmentation on the S3DIS \cite{s3dis} Area 5 benchmark, from which we can find that Retro-FPN can significantly improve the segmentation performance of the backbone networks. Particularly, we achieve the best performance by integrating Retro-FPN with Point Transformer \cite{zhao2021point} and yield a state-of-the-art record of 73.0 in terms of mIoU. Additionally, by integrating with KPConv \textit{deform}, Retro-FPN is able to improve the overall performance by 3.6 in terms of mIoU. It is worth noting that RFCR \cite{gong2021omni} also adopts KPConv \textit{deform} as backbone, which improves performance (1.6 on mIoU) by enhancing the feature pyramid on region level semantic information. Compared with RFCR, Retro-FPN can better stimulate the potential of the backbone network (3.6 versus 1.6 in terms of mIoU improvements over KPConv \textit{deform}), this should be credited to the retrospective refinement on point-level semantic features.
Furthermore, by assembling with KPConv \textit{rigid} \cite{thomas2019kpconv}, Retro-FPN is able to significantly raise mIoU by 4.3. In addition, Retro-FPN can also improve the voxel-based MinkowskiNet \cite{Choy_2019_CVPR} by 4.1 in terms of mIoU. Note that the intermediate layers of voxel-based methods lack precise per-point information due to the convolution, our Retro-FPN can complement the drawback and explicitly extracts point-level semantic information from voxel features.

\noindent\textbf{S3DIS 6-fold.} In Table \ref{table:s3dis_6fold}, we show the quantitative results of the 6-fold cross validation on S3DIS \cite{s3dis} dataset. From Table \ref{table:s3dis_6fold}, we can find that Retro-FPN can significantly improve over Point Transformer by 3.8 absolute percentage points. The result indicates that although the Point Transformer is a strong baseline, it still suffers from the information loss of implicit region features and Retro-FPN can still improve its performance robustly.

\noindent\textbf{ScanNet V2.} In table \ref{table:scannet}, we evaluate the performance of Retro-FPN on ScanNet v2 \cite{dai2017scannet} dataset. We follow the same practice of \cite{hu2021bidirectional, Choy_2019_CVPR, nekrasov2021mix3d} and adopt MinkowskiNet as the backbone to conduct experiments under voxel size 2cm and 5cm. As shown in Table \ref{table:scannet}, Retro-FPN is able to improve the segmentation performance under various voxel sizes, where Retro-FPN raises mIoU by 2.4 and 1.9 under voxel size of 5cm and 2cm, respectively. Also, Retro-FPN improves the result of MinkowskiNet on the test set to 74.4. Moreover, by integrating with the state-of-the-art Point Transformer V2 \cite{wu2022pointv2}, Retro-FPN can still improve the mIoU on validation set by 0.6, which justifies the effectiveness of Retro-FPN.

\noindent\textbf{SemanticKITTI}. Besides indoor datasets, we also integrate Retro-FPN with MinkowskiNet \cite{Choy_2019_CVPR} and evaluate its performance on the SemanticKITTI benchmark. Following the same experimental settings of SPVNAS \cite{tang2020searching}, we report the mIoU on both the validation and test sets. From Table \ref{table:semantickitti}, we can find that Retro-FPN can improve the mIoU by 3.5 and 3.9 on the validation and test sets, respectively. The results on both the indoor and outdoor benchmarks can well demonstrate the effectiveness of Retro-FPN.

\begin{table}
  \caption{Quantitative results on the SemanticKITTI \cite{semantickitti} benchmark. We report the mIoU on the validation and test sets. * means that rotation augmentation on the test set is applied.}
      \label{table:semantickitti}
      \centering
  \begin{tabular}{l|cc}
  \toprule
  Method & Val & Test\\
       \midrule
       KPConv \cite{thomas2019kpconv} & - & 58.8 \\
       FusionNet \cite{fusionnet} * & - & 61.3\\
       KPRNet \cite{kochanov2020kprnet} & - & 63.1\\
       JS3C-Net \cite{yan2021sparse} &  & 66.0\\
       SPVNAS \cite{tang2020searching} * & 64.7 & 66.4 \\
       Cylinder3D \cite{zhu2021cylindrical} * & - & 68.9 \\
       RPVNet \cite{RPVNet} & - & 70.3 \\
       $(\mathrm{AF})^2$-S3Net \cite{AF2S3NET} & - & 70.8\\
       PVKD \cite{PVKD} * & - & 71.2 \\
       2DPASS \cite{yan20222dpass} * & - & 72.9 \\
       \midrule
       MinkowskiNet \cite{Choy_2019_CVPR} * & 61.9 & 64.1 \\
       MinkowskiNet + Retro-FPN * & \textcolor{red}{65.4} & \textcolor{red}{68.0}\\
       \bottomrule
  \end{tabular}
\end{table}

\begin{figure}[!t]
   \centering
   \includegraphics[width=\linewidth]{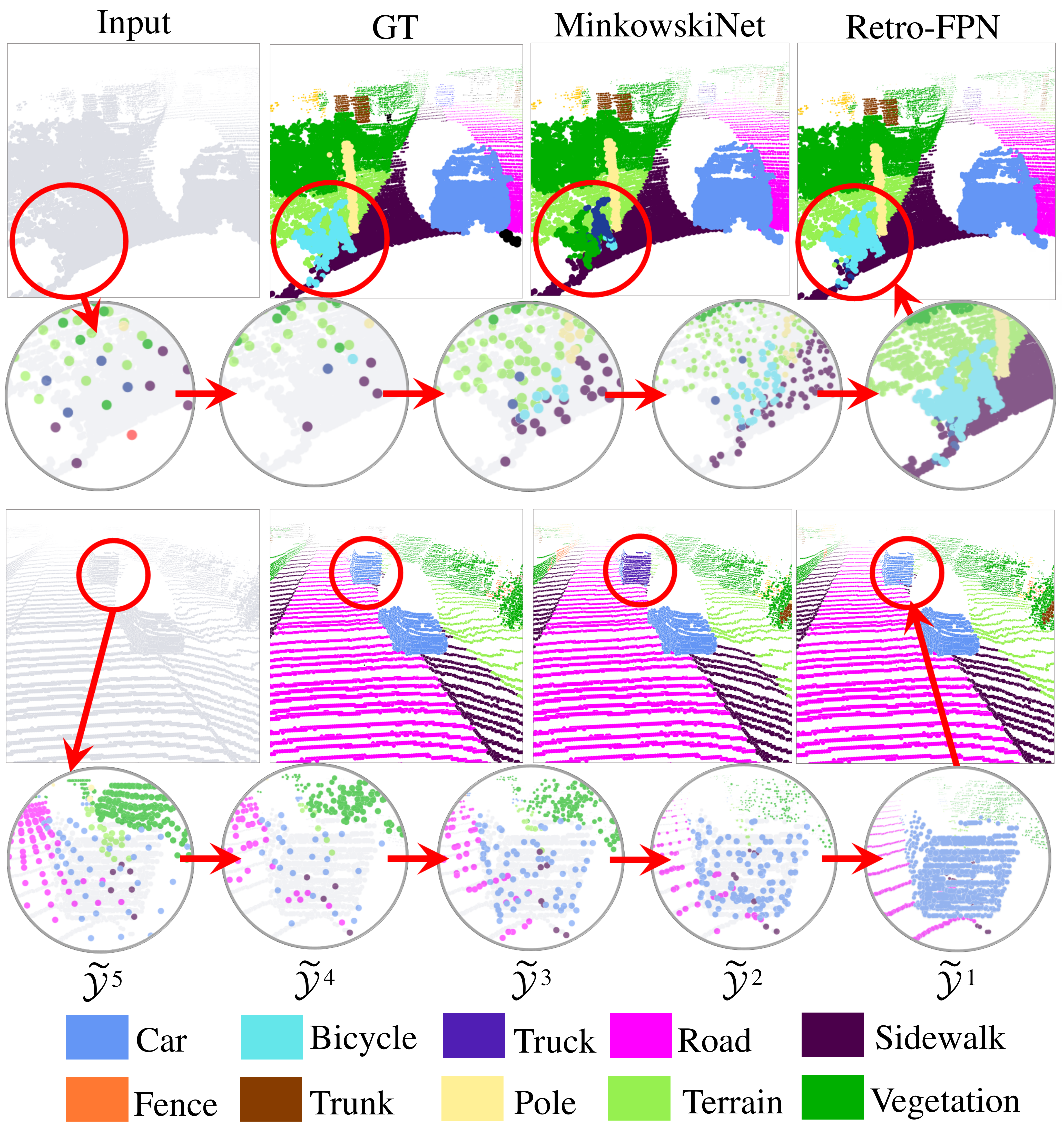}
   \caption{Visualization results of Retro-FPN and the improvements over the backbone networks. The circular areas highlighted in blue visualize the refining process of the improved areas.}
   \label{fig:visualization}
\end{figure}

\subsection{Qualitative results}
\label{section:qualitative}
In Figure \ref{fig:visualization}, we give the visualization results of Retro-FPN and the qualitative improvements over the backbone (MinkowskiNet \cite{Choy_2019_CVPR}). Moreover, we also visualize the refining process of semantic labels in each layer, which is highlighted in black circles. The visual results show that Retro-FPN can help to improve segmentation in challenging areas, such as the bicycle in the first example and the car in the second example. The improved ability of perceiving small objects should be credited to the retrospective refinement on point-level semantic information.

\begin{table}
  \caption{Effect of each part in retro-transformer. \textbf{HS}: hierarchical supervision \textbf{Cross-att}: local cross-attention. \textbf{PointEmb}: learnable position embedding. \textbf{SemGate}: semantic gate unit.}
      \label{table:retro_transformer}
      \centering
  \begin{tabular}{l|cccc|c}
  \toprule
  ID & HS & Cross-att & PosEmb & SemGate & mIoU\\
       \midrule
       
  \uppercase\expandafter{\romannumeral1} & & & & & 70.4 \\
  \uppercase\expandafter{\romannumeral2} & \checkmark & & & & 70.6 \\
  \uppercase\expandafter{\romannumeral3} & \checkmark & \checkmark & & & 71.9 \\
  \uppercase\expandafter{\romannumeral4} & \checkmark & \checkmark & \checkmark & & 72.4 \\
  \uppercase\expandafter{\romannumeral5} & \checkmark & \checkmark  &  & \checkmark & 72.2 \\
  \uppercase\expandafter{\romannumeral6} & & \checkmark & \checkmark & \checkmark &  70.8 \\
  \uppercase\expandafter{\romannumeral7} & \checkmark & \checkmark & \checkmark & \checkmark &  \textbf{73.0} \\
       \bottomrule
  \end{tabular}
  \vspace{-1em}
\end{table}

\vspace{-1em}
\section{Model Analysis}

In this section, we first provide ablation study regarding each part in Retro-FPN, then we analyze the method in terms of model complexity and run-time efficiency. More model analysis is provided in the supplementary materials.

\subsection{Ablation study}
\label{section:ablation}
We analyze the effect of each part in Retro-FPN in Table \ref{table:retro_transformer}, where we typically choose Point Transformer \cite{zhao2021point} as the backbone and analyze Reto-FPN on the S3DIS \cite{s3dis} Area 5 benchmark. Note that retro-transformer consists of vanilla cross-attention (Cross-att), learnable position embedding (PosEmb) and semantic gate unit (SemGate). 

\noindent\textbf{Effect of explicit refinement.} By comparing Exp. \uppercase\expandafter{\romannumeral2}, \uppercase\expandafter{\romannumeral6} with the baseline \uppercase\expandafter{\romannumeral1}, we show that hierarchical supervision (HS) and retro-transformer is an inseparable integration, neither of them can't take effect alone. Without HS guiding per-point predictions, the retro-transformer still suffers from the ambiguous region features and cannot fully utilize the feature pyramid. Meanwhile, without retro-transformer to refine per-point semantic information, the explicit intermediate features produced by HS cannot facilitate the final prediction. Because the backbone region features with large receptive fields serve to capture multi-class information within local regions, which may not be enhanced by the per-point single class labels. 
Exp. \uppercase\expandafter{\romannumeral2} and \uppercase\expandafter{\romannumeral6} can prove the importance of explicit refinement on point-level semantic information.

\noindent\textbf{Effect of retrospective refinement.} By comparing Exp. \uppercase\expandafter{\romannumeral3} with the baseline (Exp. \uppercase\expandafter{\romannumeral1}), we can find that the Retro-FPN with the vanilla cross-attention can already improve the backbone by 1.5 in terms of mIoU, which justifies the effectiveness of retrospective refinement.

\noindent\textbf{Effect of retro-transformer.} The results of Exp. \uppercase\expandafter{\romannumeral4}, \uppercase\expandafter{\romannumeral5} and \uppercase\expandafter{\romannumeral7} indicate that both the learnable position embedding and the semantic gate unit can further improve the refining capacity upon the vanilla cross-attention. Since the local distribution of points may change dramatically, the learnable position embedding can help the local cross-attention to better capture positional relationships. And the semantic gate unit can further screen and control semantic information refinement. Moreover, the combination of PosEmb and SemGate improves mIoU by 1.1 over the vanilla cross-attention, which further validates the design of retro-transformer.

\begin{table}[ht]

\caption{Run-time model complexity compared with backbones.}
\centering
    \resizebox{\linewidth}{!}{\begin{tabular}{l|l|c|c|c|c}
    \toprule
    Dataset & Method  & Params (M) & Latency (s)  & Mem (G) & mIoU\\
     \midrule
    \multirow{8}{*}{S3DIS \cite{s3dis}} & MinkowskiNet \cite{Choy_2019_CVPR} & 15.49 & 4.44  & 3.07 & 65.4\\
    & +Retro-FPN & 15.57 & 5.58 & 4.42 & 69.5\\
    \cmidrule{2-6}
    & KPConv \textit{rigid} \cite{thomas2019kpconv} & 24.38 & 3.81 & 4.88 & 65.4\\
    & +Retro-FPN & 24.65 & 4.64 & 5.48  & 69.7\\
    \cmidrule{2-6}
    & KPConv \textit{deform} \cite{thomas2019kpconv} & 25.59 & 4.96 & 5.69 & 67.1\\
    & +Retro-FPN & 25.86  & 6.32 & 6.71 & 70.7\\
    \cmidrule{2-6}
    & Point Transformer \cite{zhao2021point} & 7.77  & 54.05 & 6.78 & 70.4\\
    & +Retro-FPN & 7.86 & 55.16 & 7.45 & 73.0\\ \midrule
    
    \multirow{4}{*}{ScanNet \cite{dai2017scannet}} & MinkowskiNet \cite{Choy_2019_CVPR} & 15.49 & 3.14 & 3.81 & 68.0\\
    & +Retro-FPN & 15.57 & 4.06 & 5.16 & 70.8\\
    \cmidrule{2-6}
    & PTV2 \cite{wu2022pointv2} & 11.32 & 20.35 & 14.75 & 75.4\\
    & +Retro-FPN & 11.52 & 23.40 & 17.71 & 76.0\\
    
    \midrule
    \multirow{2}{*}{SemanticKITTI \cite{semantickitti}} & MinkowskiNet \cite{Choy_2019_CVPR} & 21.73 & 6.82 & 3.52 & 63.1\\
    & +Retro-FPN & 21.81 & 8.84 & 4.57 & 68.0\\
    \bottomrule
    
  \end{tabular}}
  \label{table:complexity}
\end{table}

\subsection{Model Complexity}
\label{section:model_complexity}
We analyze the model complexity of Retro-FPN in Table \ref{table:complexity}, which is evaluated in terms of parameter number, inference latency and training memory consumption (Mem). To have a fair comparison, we keep the testing settings the same as backbone networks. The inference latency is computed by randomly selecting a scene/scan and summing the inference time of 100 forward passes. For training memory consumption, we set the batch size of all methods to one and record the maximal memory consumption required during one training epoch. The results in Table \ref{table:complexity} show that Retro-FPN leads to negligible extra parameters, ranging from 0.08M to 0.27M. Particularly, for MinkowskiNet on the SemanticKITTI dataset, the increased parameter number (0.08M) is only 0.37\% of the backbone network (21.73M). Meanwhile, Retro-FPN leads to consistent computation cost across all backbones, ranging from 0.83s to 2.02s. For lightweight backbones (MinkowskiNet and KPConv), Retro-FPN leads to 20\%-30\% extra computation overhead. For Point Transformer backbone, Retro-FPN introduces marginal computation cost of 1.11s, which is 2.1\% of Point Transformer (54.05s in terms of inference time). As for training memory consumption, the extra memory required by Retro-FPN is also consistent across different backbones (except for Point Transformer V2), which ranges from 0.67G to 1.35G. For the Point Transformer V2 baseline, the extra 2.95G memory used is 20.1\% of the backbone, which is controlled in a reasonable range.
In summary, the extra parameters are negligible. The extra computation cost and memory consumption can be effectively controlled. Since Retro-FPN can be conveniently integrated with existing backbones, it provides a valuable trade-off among time and better performance.


\section{Conclusions and Limitations}
We present Retro-FPN to improve per-point semantic feature prediction for 3D point clouds, which can fully exploit the feature pyramid and models the feature propagation as an explicit and retrospective refining process on point-level semantic information. By further introducing a retro-transformer in each pyramid layer, Retro-FPN can effectively extract and refine semantic information from all pyramid levels to the final prediction layer. We integrate Retro-FPN with three prevailing backbones and conduct experiments on widely used benchmarks. Experimental results demonstrate that Retro-FPN can significantly improve segmentation performance over state-of-the-art methods.


The primary limitation of Retro-FPN is that the retro-transformer relies on the $\mathrm{K}$-NN search to capture local semantic contexts. Since the point distribution of point clouds may vary dramatically in different local regions, a fixed number of nearest neighbors may fail to provide informative contextual information for refinement, especially in dense and complex areas. Meanwhile, a large number of $\mathrm{K}$-NN search will also lead to more computation cost. Therefore, a promising future direction is explore flexible neighbor searching strategy, in order to capture more accurate semantic contexts and further bring down computation cost.


{\small
\bibliographystyle{ieee_fullname}
\bibliography{ref}
}

\end{document}